\pdfoutput=1
\def\RSSAnonymous{0}
\providecommand{\RSSAnonymous}{1}
\documentclass[conference]{IEEEtran}
\usepackage{times}
\usepackage[numbers,sort]{natbib}
\usepackage{multicol}
\usepackage{amsmath,amssymb,stmaryrd}
\usepackage{graphicx}
\usepackage{xcolor}
\usepackage{pgfplots}
\usepgfplotslibrary{fillbetween}
\usetikzlibrary{arrows.meta,calc,svg.path}
\pgfplotsset{compat=1.18}
\usepackage{fix-cm}

\definecolor{rssSCI}{RGB}{68,120,171}
\definecolor{rssSCP}{RGB}{160,45,60}
\colorlet{rssContactTotal}{rssSCI}
\colorlet{rssContactPair}{rssSCP}
\newsavebox{\rssPlotBox}
\newlength{\rssPlotAxisWidth}

\pgfplotsset{
  rss axis/.style={
    scale only axis,
    width=\rssPlotAxisWidth,
    height=3.6cm,
    axis lines=box,
    axis line style={line width=.3pt},
    tick style={line width=.4pt},
    tick align=inside,
    major tick length=3pt,
    minor tick length=1.5pt,
    ticklabel style={font=\small},
    label style={font=\small},
    xlabel style={inner sep=2pt},
    ylabel style={inner sep=2pt},
    scaled ticks=false,
    grid=none,
  },
  rss sci/.style={color=rssSCI,line width=.8pt,mark=square*,
    mark size=1.7pt,mark options={solid,fill=white,line width=.5pt}},
  rss scp/.style={color=rssSCP,line width=.8pt,mark=*,
    mark size=1.7pt,mark options={solid,fill=white,line width=.5pt}},
}

\usepackage{framed}
\usepackage{algorithm}
\usepackage{algpseudocode}
\usepackage[bookmarks=true]{hyperref}
\usepackage[capitalize,noabbrev,nameinlink]{cleveref}
\crefname{subsection}{Section}{Sections}
\Crefname{subsection}{Section}{Sections}
\crefname{subsubsection}{Section}{Sections}
\Crefname{subsubsection}{Section}{Sections}
\newtheorem{proposition}{Proposition}

\crefformat{supplement}{#2Supplementary Material#3}
\Crefformat{supplement}{#2Supplementary Material#3}

\newcounter{experiment}
\newenvironment{experiment}[1][]{%
  \par\addvspace{0.7\baselineskip}%
  \refstepcounter{experiment}%
  \noindent\textbf{Experiment \theexperiment\ (#1).}\ \ignorespaces
}{\par\addvspace{0.4\baselineskip}}
\crefname{experiment}{Experiment}{Experiments}
\Crefname{experiment}{Experiment}{Experiments}
\usepackage[acronym]{glossaries}
\glsdisablehyper
\newacronym{sci}{SCI}{sequential contact insertion}
\newacronym{scp}{SCP}{sequential convex programming} 
\providecommand{\norm}[1]{\lVert #1 \rVert}
\providecommand{\inner}[1]{\langle #1 \rangle}

\providecommand{\dd}{\mathop{}\!\mathrm{d}}
\providecommand{\jmp}[1]{\llbracket #1 \rrbracket} 
\title{Contact-based optimization for collision-free motion planning for a million robots}
\hypersetup{pdftitle={Contact-based optimization for collision-free motion planning for a million robots},pdfauthor={}}
\ifnum\RSSAnonymous=1
  \author{Author Names Omitted for Anonymous Review. Paper-ID 97}
\else
  \hypersetup{pdfauthor={Jan Kamm, Antonio Terpin, Raffaello D'Andrea}}
  \author{Jan Kamm\textsuperscript{*}, Antonio Terpin\textsuperscript{*},
    Raffaello~D'Andrea%
    \thanks{\textsuperscript{*} These authors contributed equally to this work.\newline
      Corresponding author: Jan Kamm
      (\href{mailto:jakamm@ethz.ch}{jakamm@ethz.ch}).
      All authors are with the Institute for Dynamic Systems and Control, ETH Zurich, Zurich, Switzerland.
      }}
\fi
\IEEEoverridecommandlockouts
\makeatletter
\IEEEaftertitletext{%
  \begin{minipage}{\textwidth}
    \def\@captype{figure}
    \centering
    \begingroup

\newcommand{\rssMathFont}{\rmfamily\fontsize{10}{12}\selectfont}
\rssMathFont
\definecolor{initbg}{RGB}{250,248,235}
\definecolor{initborder}{RGB}{205,187,68}
\definecolor{contactbg}{RGB}{240,245,250}
\definecolor{contactborder}{RGB}{68,120,171}
\definecolor{accentred}{RGB}{160,45,60}
\definecolor{janzGreen}{RGB}{240,248,245}
\raisebox{0sp}[11432821sp][0sp]{\includegraphics[width=33750700sp,height=11432821sp]{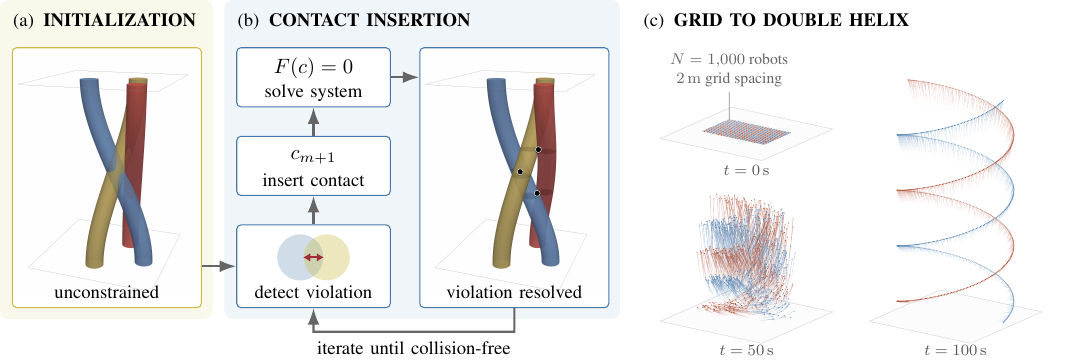}}%
\endgroup{} \par
    \caption{\Glsentrylong{sci}.
      (a) Initialization with unconstrained trajectories.
      (b) Algorithm loop: detect separation violations, insert a contact,
      and solve the algebraic system until collision-free.
      Colored tubes in (a,b) show planar robot trajectories in space-time, with time as
      the vertical axis; black dots mark contacts.
      (c) Grid-to-double-helix transition for $N=1000$ robots on a $40\times25$
      grid (\cref{exp:million-robots}): snapshots at $t=0,50,100\,\mathrm{s}$;
      trails show the preceding $20\,\mathrm{s}$.
      Blue and rust identify the two target strands.}
    \label{fig:header}
    \ifnum\RSSAnonymous=0
      \par\vspace{8pt}
      {\small\urlstyle{tt}\hypersetup{hidelinks}%
        \raisebox{-148891sp}[414311sp][148891sp]{\includegraphics[width=576720sp,height=563202sp]{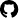}}\hspace{0.5em}%
        \url{https://github.com/jankammeth/sci}%
        \hspace{2em}%
\raisebox{\dimexpr0.5ex-0.5\height\relax}{\resizebox{!}{10pt}{%
\raisebox{0sp}[3801088sp][0sp]{\includegraphics[width=4313058sp,height=3801088sp]{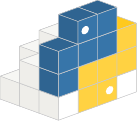}}}}\hspace{0.5em}%
        \url{https://pypi.org/project/sci-solver/}%
        \par}
    \else
      \par\vspace{6pt}
    \fi
  \end{minipage}\par
}
\makeatother
\begin{document}
\maketitle
\glsresetall

\begin{abstract}
Multi-agent robotic systems have attracted growing
interest in recent years, with applications ranging from coordinated drone shows
to autonomous operations on land, in the air, and on water. At the planning
level, robot motion is often modeled using double-integrator dynamics
(with low-level controllers handling the physical dynamics during execution)
with a
minimum-energy objective subject to collision-avoidance constraints and 
limits on position, velocity, and acceleration. 
The complexity of the problem resides in the constraints, which couple the 
trajectories and make the optimization problem nonconvex. 
Existing numerical approaches typically discretize and optimize entire trajectories, which becomes expensive for large fleets. We exploit the structure of optimal trajectories by organizing optimization around contacts, at which robots reach the minimum allowed separation. We formulate a coupled algebraic system in contact variables and reconstruct trajectories analytically between contacts. Our algorithm, \gls{sci}, starts from unconstrained trajectories, inserts contacts at separation violations, and jointly adjusts their times and states. Numerical experiments demonstrate planning for one million robots in minutes and thousands of robots in fractions of a second. On problems feasible for a fixed-grid \glsentrylong{scp} baseline, \gls{sci} achieves median speedups of several orders of magnitude at matching solution accuracy.
\end{abstract}

\IEEEpeerreviewmaketitle
\glsresetall
\section{Introduction}
\label{sec:introduction}
Multi-agent robotic systems have attracted growing interest in recent
years, with applications ranging from coordinated drone shows to autonomous
operations on land \citep{wurman_coordinating_2008}, in the air
\citep{augugliaro_generation_2012}, and on water \citep{kamm_floating_2026}.
Planning collision-free motion for a robot fleet is difficult because each
robot's trajectory constrains the motion of the others. We consider transitions
between prescribed initial and final states over a fixed time horizon with
double-integrator dynamics, a squared-acceleration objective,
collision-avoidance and box constraints on position, velocity, and acceleration. This model is
commonly used in the literature for fleet motion planning
\citep{augugliaro_generation_2012,luis_trajectory_2019,kamm_floating_2026,beaver_assignment_2020,beaver_energy_2021}, typically paired with
low-level controllers that handle the physical dynamics during
execution.
The complexity of the problem resides in the pairwise separation constraints that make the
joint optimization problem nonconvex. The problem is therefore often solved
numerically \citep{shome_drrt_2020,honig_trajectory_2018,vanloock_spline_2015}, with
\gls{scp} as a popular choice \citep{augugliaro_generation_2012,schulman_trajectories_2013,hauser_semi_infinite_2021,zhang_subdivision_2024,elango_continuous_2025}.

A different route to efficient planning, and the one explored in this paper,
is to exploit the structure of optimal
trajectories. State-constrained optimal control supplies conditions
for unconstrained and constrained trajectory pieces and for matching them
\citep{hartl1995,bonnans_hermant_2007}.
Shooting methods use these conditions to construct a finite system of
equations for these pieces and the times at which they meet
\citep{bonnans_hermant_2007}.
For collision avoidance, \citet{beaver_energy_2021} derive such matching
conditions for one agent relative to a prescribed reference trajectory.
For single-robot planning around fixed obstacles,
\citet*{beaver_graph_2023} solve trajectory equations within a search over
obstacle-vertex sequences.
For fleets, \citet{beaver_assignment_2020} solve a square algebraic system
for each robot's trajectory while holding higher-priority trajectories fixed.
Finally, \citet{kamm_characterization_2026} completely characterize optimal two-robot motion with zero initial and final velocities
for the squared-acceleration objective considered here.
We exploit this characterization to construct solutions for larger fleets.

\noindent\textbf{In this work.}
We formulate fleet planning in terms of pairwise contacts, at which robots
reach the prescribed minimum separation. The stationarity conditions yield
a coupled algebraic system for the contact times and states, from which
the trajectories between contacts are reconstructed analytically.
The contact graph decomposes this system into independent groups of
interacting robots.
Our approach accounts for position, velocity,
and acceleration box constraints.
\Gls{sci} starts from independent robot trajectories, inserts contacts
at separation violations, and jointly adjusts their times and states.
\Cref{fig:header} illustrates the algorithm and a grid-to-double-helix
transfer for $N=1000$ robots. We scale \gls{sci} to $N=10^6$ robots,
which is, to our knowledge, unprecedented for joint continuous-time
fleet trajectory optimization.
For fleet sizes amenable to existing methods, \gls{sci} obtains median speedups of several orders
of magnitude over an \gls{scp} baseline at matching accuracy.
\section{Problem setup}
\label{sec:problem_setup}

\subsection{Fleet model}

Consider $N$ robots in $\mathbb{R}^d$, $d\geq2$, and a fixed time horizon
$T>0$. Robot $i$ has position $x_i(t)$ and prescribed initial and terminal
states $(x_{i,0},\dot{x}_{i,0})$ and $(x_{i,T},\dot{x}_{i,T})$.
Let $R>0$ be the prescribed minimum separation distance between any two
robots.
Throughout, we assume that the endpoint configurations are strictly separated:
$\norm{x_i(t)-x_j(t)}>R$ for $i<j$ and $t\in\{0,T\}$.
Componentwise position, velocity, and acceleration limits are prescribed by
constant, nonempty boxes
$\mathcal B_i^{(k)}=\{z\in\mathbb R^d:
\underline b_i^{(k)}\leq z\leq\overline b_i^{(k)}\}$,
$k=0,1,2$, respectively; the bounds $\underline b_i^{(k)}$ and $\overline b_i^{(k)}$ are vectors in $\bar{\mathbb R}^d, \bar{\mathbb R} = \mathbb{R} \cup\{-\infty, \infty\}$,
and the inequalities are understood componentwise.

We use double-integrator dynamics and optimize position
trajectories with absolutely continuous velocities and square-integrable
accelerations. We seek the fleet trajectory $x=(x_1,\ldots,x_N)$ that
minimizes total squared acceleration subject to endpoint states,
collision avoidance, and box bounds:
\begin{equation}
  \begin{aligned}
    \min_x \quad & J[x]=\sum_{i=1}^{N}\int_0^T
      \norm{\ddot{x}_i(t)}^2\dd t \\
    \text{s.t.} \quad
    & (x_i(0),\dot{x}_i(0))=(x_{i,0},\dot{x}_{i,0}), \\
    & (x_i(T),\dot{x}_i(T))=(x_{i,T},\dot{x}_{i,T}), \\
    & \norm{x_i(t)-x_j(t)}\geq R, \\
    & x_i(t)\in\mathcal B_i^{(0)},
    \quad
      \dot{x}_i(t)\in\mathcal B_i^{(1)}, 
      \quad
      \ddot{x}_i(t)\in\mathcal B_i^{(2)}.
  \end{aligned}
  \label{eq:fleet-problem}
\end{equation}
Endpoint conditions and box bounds apply to every robot, and separation
constraints apply to every pair $i<j$.
Separation, position, and velocity bounds hold for all $t\in[0,T]$;
acceleration bounds hold almost everywhere.

\subsection{Barycentric decomposition}
\label{sec:barycentric-reduction}

Define the barycenter $\bar{x}=N^{-1}\sum_{i=1}^N x_i$ and centered
positions $\tilde{x}_i=x_i-\bar{x}$. The pairwise relative positions
$r_{ij}=x_i-x_j=\tilde{x}_i-\tilde{x}_j$ determine separation, while
the acceleration cost splits as
$\sum_{i=1}^N\norm{\ddot{x}_i(t)}^2
=N\norm{\ddot{\bar{x}}(t)}^2
+\sum_{i=1}^N\norm{\ddot{\tilde{x}}_i(t)}^2$.
The boxes constrain $x_i=\bar{x}+\tilde{x}_i$ and its derivatives,
coupling barycentric and centered motion.
When all boxes equal $\mathbb R^d$, the barycenter minimizes its own
squared-acceleration cost with fixed endpoint states. At every local
minimum of this specialization, it is the cubic Hermite interpolant
$\bar{x}^{\mathrm H}$ of the mean endpoint positions and velocities
\citep[Section~2.5]{egerstedt_martin_2010}.

\subsection{Necessary conditions}
\label{sec:contact-laws}

For each pair, consider the times at which $\norm{r_{ij}(t)}=R$.
An isolated such time is an isolated contact; an interval of
positive length represents sustained contact.
We represent the dual multipliers for continuous collision, position, and
velocity constraints by finite Radon measures.
Collision multipliers satisfy $\mu_{ji}=\mu_{ij}\geq0$
and are supported at contact times.
For position ($k=0$) and velocity ($k=1$) bounds, write
$\nu_i^{(k)}=\nu_i^{(k),-}-\nu_i^{(k),+}$, where the nonnegative
component measures are supported on the corresponding active lower and
upper bounds by complementary slackness
\citep{hartl1995}.
At a local minimum admitting normal multipliers, stationarity reads
\begin{equation}
  \ddot{x}_i=\operatorname*{proj}_{\mathcal B_i^{(2)}}w_i,\quad
  D_t^2w_i=\sum_{j\ne i}r_{ij}\,\mu_{ij}
    +\tfrac12\bigl(\nu_i^{(0)}-D_t^1\nu_i^{(1)}\bigr),
  \label{eq:bounded-balance}
\end{equation}
where $w_i$ is an auxiliary momentum and $D_t^k$ denotes the
distributional derivative on $(0,T)$.
The projection clips each component of $w_i$ to the corresponding
acceleration interval~\citep{bauschke_projection_2019}.

\paragraph*{Unrestricted boxes}
When all boxes equal $\mathbb R^d$, the state-bound multipliers vanish
and $w_i=\ddot{x}_i$, reducing \eqref{eq:bounded-balance} to
\begin{equation}
  D_t^4x_i=\sum_{j\ne i}r_{ij}\,\mu_{ij}.
  \label{eq:fleet-balance}
\end{equation}
This extends the two-robot contact conditions of
\citet{kamm_characterization_2026} to fleets.
The sum couples all encounters involving a robot.
On an interval without contacts involving robot $i$, the right-hand
side vanishes and $x_i^{(4)}=0$, so its motion is cubic.

The acceleration is continuous, and the jerk $x_i^{(3)}$, taken as
the right-continuous representative of $D_t^3x_i$, has one-sided limits.
Write $\jmp{f}_t=f(t^+)-f(t^-)$ for a jump between these limits.
Then, at an isolated contact $t_\ell$ involving only the pair $p<q$, define
the contact weight $m_\ell=\mu_{pq}(\{t_\ell\})\geq0$ and set
$\epsilon_p=1$, $\epsilon_q=-1$, and $\epsilon_i=0$ for all other robots.
The adjoining pieces satisfy $\jmp{\ddot{x}_i}_{t_\ell}=0$ and
$\jmp{x_i^{(3)}}_{t_\ell}=m_\ell\epsilon_i r_{pq}(t_\ell)$.
Thus only the contacting robots have jerk jumps, equal and opposite along
their relative position.
At an interior contact, separation reaches a local minimum, giving
$\inner{r_{pq}(t),\dot r_{pq}(t)}=0$. 
\section{Contact-based trajectory reconstruction}
\label{sec:contacts}

\subsection{Finite multiplier representation}

We approximate the collision multipliers by a finite list of point contacts.
Contact $\ell$ has a prescribed pair $(p_\ell,q_\ell)$, time $t_\ell$,
and weight $m_\ell\geq0$:
$\mu_{pq}=\sum_{\ell:\,(p_\ell,q_\ell)=(p,q)}
m_\ell\delta_{t_\ell}$.
Here $\delta_{t_\ell}$ is the unit Dirac measure at $t_\ell$.
We approximate the dual multipliers associated with constraints on position
by atoms, and those associated with constraints on velocity by atoms and
constant interval densities.

For a fixed contact list, the stationarity equations can be solved
independently on each connected component of the contact graph.
This graph has one vertex per robot and an edge for each listed pair.
We call its connected components the \emph{contact components}.
Box bounds act on individual robots, so they preserve this decomposition.

To start, take all boxes to be
$\mathbb R^d$.
Within a component of $N$ robots, use local labels $1,\ldots,N$ and
distinct contact times $0<t_1<\cdots<t_m<T$.
We use robot $1$ as a reference and store the $N-1$ relative positions
$r_{1j}$, $j=2,\ldots,N$.
Together with the fixed cubic Hermite barycenter, these determine every
robot's position. Set $r_{11}=0$. Then
$r_{pq}=r_{1q}-r_{1p}$,
$x_1=\bar{x}^{\mathrm H}+N^{-1}\sum_{j=2}^N r_{1j}$, and
$x_j=x_1-r_{1j}$ for $j\geq2$.

A contact between robots $p<q$ at time $t_\ell$ has relative state
$r_{pq}(t_\ell)=Rn_\ell$ and
$\dot{r}_\ell=\dot{r}_{pq}(t_\ell)$, with
$\norm{n_\ell}^2=1$ and $\inner{n_\ell,\dot{r}_\ell}=0$.

Each contact vector $c_\ell$ stores $t_\ell$, $n_\ell$, $\dot r_\ell$,
$m_\ell$, and the states of $N-2$ other relative coordinates.
We omit coordinate $q$, since the contact state determines it from
coordinate $p$:
\begin{equation*}
  c_\ell=\bigl(t_\ell,n_\ell,\dot{r}_\ell,m_\ell,
    (r_{1j}(t_\ell),\dot{r}_{1j}(t_\ell))_
      {j\in\{2,\ldots,N\}\setminus\{q\}}\bigr).
\end{equation*}
The omitted coordinate follows from
$r_{1q}(t_\ell)=r_{1p}(t_\ell)+Rn_\ell$ and
$\dot r_{1q}(t_\ell)=\dot r_{1p}(t_\ell)+\dot r_\ell$.
Between consecutive contact times, each $r_{1j}$ is reconstructed by
cubic Hermite interpolation
\citep{kamm_characterization_2026}.
The first and last pieces use the prescribed endpoint states.
Thus $c=(c_1,\ldots,c_m)$ determines the component trajectory.

\subsection{Matching equations}
\label{sec:matching-system}

In the relative coordinates $r_{1j}$, $j=2,\ldots,N$, the conditions for
acceleration continuity and jerk jumps from \cref{sec:contact-laws} become
$\jmp{\ddot{r}_{1j}}_{t_\ell}=0$ and
$\jmp{r_{1j}^{(3)}}_{t_\ell}=m_\ell R(\epsilon_1-\epsilon_j)n_\ell$.
When $\epsilon_1-\epsilon_j=0$, continuity of acceleration and jerk
makes the adjoining pieces of $r_{1j}$ the same cubic.

Appending normalization and tangency to these conditions gives $F(c)=0$, a system of
$m[2d(N-1)+2]$ scalar equations in as many unknowns.
The equations at $t_\ell$ involve only the two adjacent cubic pieces,
hence only $c_\ell$, its neighboring contacts, and the fixed endpoint
data. Accordingly, the Jacobian of $F$ is block tridiagonal, with blocks
of size $2d(N-1)+2$.

Consider stationary trajectories admitting multipliers with finite support
and at most one pair with a nonzero multiplier per instant in each
contact component.

 \begin{proposition}[Contact representation \unskip]\MakeLinkTarget{proposition}\label{prop:contact-representation}
The continuously feasible reconstructions of roots of $F(c)=0$ with
$m_\ell\geq0$ are exactly the stationary fleet trajectories in this class. \end{proposition}  
\begin{IEEEproof}
  For a continuously feasible reconstruction of a root,
the reconstruction formulas and matching conditions give
continuous acceleration and the jerk jumps of \cref{sec:contact-laws}.
Its fourth distributional derivative is therefore the atomic right-hand
side of \eqref{eq:fleet-balance}, with nonnegative weights and support
on contacts.
Conversely, summing \eqref{eq:fleet-balance} over each component cancels
the multiplier contributions, fixing its barycenter to the prescribed Hermite cubic.
Between atoms, trajectories are cubic and recovered by Hermite interpolation.
At atoms, feasibility gives tangency and the stationarity equation gives
the matching conditions.
Reading off contact states and weights yields a root reconstructing the trajectory. 
\end{IEEEproof}

\paragraph*{Sustained contact}
If the relative position changes during sustained contact, the motion
cannot be represented by finitely many cubic pieces. Approximating it
with vanishing error requires growing contact counts.
\Gls{sci} approximates sustained contact by clusters of point contacts,
checking separation between them to a prescribed tolerance.

\paragraph*{Box bounds}
With the finite multiplier representation, \eqref{eq:bounded-balance}
gives piecewise-affine momenta $w_i$.
We project $w_i$ onto the acceleration box, then integrate once to obtain
velocity and again to obtain position.
Clipping an affine function produces constant and affine pieces, so the
resulting trajectories remain piecewise polynomial. 
\section{Algorithm: \texorpdfstring{\glsentryshort{sci}}{\glsentryshort{sci}}}
\label{sec:algorithm}
\label{sec:insertion-loop}

In \cref{sec:contacts}, we reconstructed trajectories from a prescribed
contact list. To find this list, we start from unconstrained trajectories
and insert contacts where separation fails.
We then solve the matching equations to adjust the contact times and states
jointly and reconstruct the trajectories.
If a contact has a negative weight, we remove it and solve again for the
remaining contacts before checking separation.

In \cref{alg:sci}, we use absolute tolerances
$\varepsilon_F$ for matching and endpoint states, $\varepsilon_m$ for
multiplier nonnegativity, and $\varepsilon_R$ for separation.
We return \texttt{unresolved} if matching or insertion fails or the time
budget is exhausted.

\begin{algorithm}[!ht]
  \caption{\gls{sci} without box bounds}
  \label{alg:sci}
  \renewcommand{\algorithmicrequire}{\textbf{Input:}}
  \renewcommand{\algorithmicensure}{\textbf{Output:}}
  \begin{algorithmic}[1]
    \Require Endpoint states, $T$, $R$, $\varepsilon_F$, $\varepsilon_m$, $\varepsilon_R$
    \Ensure $(c,s)$, $s\in\{\texttt{success},\texttt{unresolved}\}$
    \State $c\gets()$
    \Loop
      \Loop
        \State $\mathcal G\gets\Call{Groups}{c}$
        \State $(c,s)\gets\Call{Match}{c,\mathcal G}$
        \State \textbf{if} $s=\texttt{unresolved}$ \textbf{then} \Return $(c,s)$
        \State \textbf{if} $c=()$ \textbf{or} $\min_\ell m_\ell\geq-\varepsilon_m$ \textbf{then break}
        \State $\ell\gets\operatorname*{arg\,min}_j m_j$
        \State $c\gets(c_j)_{j\ne\ell}$
      \EndLoop
      \State $(p,q,t_*,\varrho)\gets\Call{Closest}{c}$
      \State \textbf{if} $\varrho\geq R-\varepsilon_R$ \textbf{then} \Return $(c,\texttt{success})$
      \State $(c,s)\gets\Call{Insert}{c,p,q,t_*}$
      \State \textbf{if} $s=\texttt{unresolved}$ \textbf{then} \Return $(c,s)$
    \EndLoop
  \end{algorithmic}
\end{algorithm}

\paragraph*{\normalfont\bfseries Grouping}
In \Call{Groups}{$c$}, we update only the contact components affected by
insertion or removal.
When a new contact links two groups, we combine their matching equations
into one system; after removal, we check whether the affected group splits
into smaller groups that we can solve independently.

\paragraph*{\normalfont\bfseries Matching}
In \Call{Match}{$c,\mathcal G$}, we first solve each group's system $F(c)=0$
by Newton iteration~\citep{dennis_schnabel_1996}, using the analytical
block-tridiagonal Jacobian from \cref{sec:matching-system} and banded linear
solves. We use backtracking to shorten each Newton step until it decreases
the residual without changing the chronological order of the contacts.
If Newton iteration fails, we use the cubic Green
function~\citep{kamm_characterization_2026} to express the intermediate
positions and velocities in terms of the contact times, normals, and weights
$(t_\ell,n_\ell,m_\ell)_{\ell=1}^m$.
We then solve this smaller system with Powell's hybrid
method~\citep{powell_hybrid_1970} and trust-region least squares.
If these attempts repeatedly converge to roots with negative weights,
we use shifted deflation~\citep{farrell_deflation_2015} to rescale the
residual near those roots and direct subsequent solves toward another root.
We accept a match only when $\norm{F(c)}_\infty\leq\varepsilon_F$ and every
endpoint position and velocity component has absolute error at most
$\varepsilon_F$.

\paragraph*{\normalfont\bfseries Detection}
In \Call{Closest}{$c$}, we find the closest pair $(p,q)$ over the full horizon
and return its separation $\varrho$ and the time $t_*$ at which it occurs.
We use Bernstein convex-hull bounds~\citep{farouki_bernstein_2012} to enclose
trajectory segments over adaptively subdivided time intervals.
We group these enclosures in bounding-volume hierarchies and compare
whole groups at once, excluding distant pairs without checking each
trajectory individually (dual-tree traversal~\citep{curtin_dual_tree_2013}).
For the remaining pairs, we find the minimum squared distance by evaluating
it at segment endpoints and at times when its derivative vanishes.

\paragraph*{\normalfont\bfseries Insertion}
In \Call{Insert}{$c,p,q,t_*$}, we initialize the new contact's state from
the current trajectory at $t_*$ and set its weight to $m_\ell=0$.
To handle symmetric and antipodal cases,
we slightly shift $t_*$ when it coincides with an existing contact time.
We process violations in order of increasing separation, adding at most
one contact involving each component before solving the matching equations
again.

\paragraph*{Position, velocity, and acceleration limits}
With box constraints, we minimize $J$ under endpoint conditions,
acceleration bounds throughout the motion, and position, velocity, and
separation inequalities at selected times.
For a fixed unit separation direction $n_\ell$, we impose
$\inner{n_\ell,r_{p_\ell q_\ell}(t_\ell)}\geq R$,
allowing the robots to move farther apart than $R$.
For velocity, we also impose interval constraints
$(x_i(b)-x_i(a))/(b-a)\in\mathcal B_i^{(1)}$, $a<b$;
the associated dual multipliers represent constant densities on $[a,b]$.
These constraints bound interval averages; we check instantaneous velocity
after each solve.
By fixing these times, intervals, and directions during each update, we
obtain a convex subproblem. We solve its dual in the multipliers associated
with endpoint conditions and the selected inequalities, requiring
nonnegative multipliers for the latter.
After each update, we check endpoint accuracy, every box bound, and
pairwise separation over the full horizon.
We add constraints where violations remain, remove constraints with
zero multipliers, and update the separation directions before solving again.
We accept the trajectory only when all checks pass. 
\section{Numerical experiments}
\label{sec:experiments}
\newcommand{\NumFleetMax}{20}
\newcommand{\NumFleetSCI}{0.009}
\newcommand{\NumFleetSCP}{106.6}
\newcommand{\NumFleetRatio}{12289}
\newcommand{\NumFleetTail}{0.1}
\newcommand{\NumKMin}{10}
\newcommand{\NumKMax}{200}
\newcommand{\NumViolationFirst}{77.5}
\newcommand{\NumViolationLast}{1.39}
\newcommand{\NumViolationTail}{4.95}
\newcommand{\NumSpeedupFirst}{5.38}
\newcommand{\NumSpeedupLast}{4756}
\newcommand{\NumCostGapFirstPercent}{0.94}
\newcommand{\NumCostGapLast}{3.6\times10^{-6}}
\newcommand{\NumCostGridGapFirstPercent}{-1.1}
\newcommand{\NumCostGridGapLastPercent}{-0.0044}
\newcommand{\NumPrioMax}{20}
\newcommand{\NumPrioSCI}{0.015}
\newcommand{\NumPrioSCP}{7.45}
\newcommand{\NumPrioRatio}{498.2}
\newcommand{\NumPrioSpeedup}{506}
\newcommand{\NumPrioPairs}{464}
\newcommand{\NumPrioPositionRMSE}{\text{unavailable}}
\newcommand{\NumPrioCostRatio}{1.000}
\newcommand{\NumPrioRankPairs}{42}
\newcommand{\NumHardware}{The archived measurements were obtained on an Apple M3 Pro.}
\newcommand{\NumFleetDesign}{50 seeds per fleet size $N\in\{2,4,\ldots,20\}$}
\newcommand{\NumKDesign}{$K\in\{10,15,25,35,50,75,100,150,200\}$}
\newcommand{\NumCongAttemptedSCI}{300}
\newcommand{\NumCongAcceptedSCI}{300}
\newcommand{\NumCongSuccessRateSCI}{100}
\newcommand{\NumCongContinuousRateSCI}{100}
\newcommand{\NumCongSCIDenseTime}{0.009}
\newcommand{\NumCongSCISparseTime}{0.002}
\newcommand{\NumCongAttemptedSCP}{300}
\newcommand{\NumCongAcceptedSCP}{98}
\newcommand{\NumCongSuccessRateSCP}{98}
\newcommand{\NumCongContinuousRateSCP}{32.7}
\newcommand{\NumCongSCPDenseTime}{11.578}
\newcommand{\NumCongSCPSparseTime}{1.938}
\newcommand{\NumCongSCIDenseContacts}{5}
\newcommand{\NumCongSCISparseContacts}{1}
\newcommand{\NumCongReturnedSCP}{294}
\newcommand{\NumCongErrorsSCP}{6}
\newcommand{\NumCongGridAcceptedSCP}{294}
\newcommand{\NumCongMedianError}{0.33}
\newcommand{\NumCongUpperError}{2.97}
\newcommand{\NumCongMaxError}{7.74}
\newcommand{\NumFleetAttempts}{50}
\newcommand{\NumFleetContactMedian}{6}
\newcommand{\NumFleetSystemSizeMedian}{65}
\newcommand{\NumFleetSuccSCPLast}{49}
\newcommand{\NumFleetAccSCPFirst}{48}
\newcommand{\NumFleetAccSCPLast}{0}
\newcommand{\NumFleetEtaSCPLast}{2.8}
\newcommand{\NumFleetAttemptsSCI}{500}
\newcommand{\NumFleetAccSCI}{500}
\newcommand{\NumFleetMultiContactSolves}{0}
\newcommand{\NumVelContactMedianRange}{4}
\newcommand{\NumVelSCIMin}{0.006}
\newcommand{\NumVelSCIMax}{0.007}
\newcommand{\NumVelSCPMin}{26.5}
\newcommand{\NumVelSCPMax}{36.7}
\newcommand{\NumAccSCIAccepted}{150}
\newcommand{\NumAccSCIAttempted}{150}
\newcommand{\NumAccSCIRuntimeMin}{0.006}
\newcommand{\NumAccSCIRuntimeMax}{0.006}
\newcommand{\NumAccContactMedian}{4}
\newcommand{\NumAccContactMedianRange}{4}
\newcommand{\NumAccCostPairs}{75}
\newcommand{\NumAccContactQOne}{3}
\newcommand{\NumAccContactQThree}{6}
\newcommand{\NumAccSCPAccepted}{191}
\newcommand{\NumAccSCPPostcheck}{286}
\newcommand{\NumAccSCPAttempted}{750}
\newcommand{\NumAccSCPTimeouts}{257}
\newcommand{\NumAccSCPErrors}{16}
\newcommand{\NumAccCostMaxDifference}{0.019}
\newcommand{\NumAccCoverageLoose}{48}
\newcommand{\NumAccContinuousRateLoose}{96}
\newcommand{\NumAccInteractingLoose}{45}
\newcommand{\NumAccSpeedupLoose}{3296}
\newcommand{\NumAccCoverageMedium}{20}
\newcommand{\NumAccContinuousRateMedium}{40}
\newcommand{\NumAccInteractingMedium}{17}
\newcommand{\NumAccSpeedupMedium}{43122}
\newcommand{\NumAccCoverageStrict}{7}
\newcommand{\NumAccContinuousRateStrict}{14}
\newcommand{\NumAccInteractingStrict}{4}
\newcommand{\NumAccSpeedupStrict}{83487}
\newcommand{\NumAblBaselineAttempts}{150}
\newcommand{\NumAblBaselineAccepted}{150}
\newcommand{\NumAblBaselineDenseAttempts}{50}
\newcommand{\NumAblBaselineDenseAccepted}{50}
\newcommand{\NumAblBaselineDenseStagnations}{0}
\newcommand{\NumAblBaselineDefaultAttempts}{50}
\newcommand{\NumAblBaselineDefaultAccepted}{50}
\newcommand{\NumAblBaselineDefaultStagnations}{0}
\newcommand{\NumAblBaselineSparseAttempts}{50}
\newcommand{\NumAblBaselineSparseAccepted}{50}
\newcommand{\NumAblBaselineSparseStagnations}{0}
\newcommand{\NumAblFdAttempts}{150}
\newcommand{\NumAblFdAccepted}{150}
\newcommand{\NumAblFdDenseAttempts}{50}
\newcommand{\NumAblFdDenseAccepted}{50}
\newcommand{\NumAblFdDenseStagnations}{0}
\newcommand{\NumAblFdDefaultAttempts}{50}
\newcommand{\NumAblFdDefaultAccepted}{50}
\newcommand{\NumAblFdDefaultStagnations}{0}
\newcommand{\NumAblFdSparseAttempts}{50}
\newcommand{\NumAblFdSparseAccepted}{50}
\newcommand{\NumAblFdSparseStagnations}{0}
\newcommand{\NumAblEarliestAttempts}{150}
\newcommand{\NumAblEarliestAccepted}{146}
\newcommand{\NumAblEarliestDenseAttempts}{50}
\newcommand{\NumAblEarliestDenseAccepted}{46}
\newcommand{\NumAblEarliestDenseStagnations}{4}
\newcommand{\NumAblEarliestDefaultAttempts}{50}
\newcommand{\NumAblEarliestDefaultAccepted}{50}
\newcommand{\NumAblEarliestDefaultStagnations}{0}
\newcommand{\NumAblEarliestSparseAttempts}{50}
\newcommand{\NumAblEarliestSparseAccepted}{50}
\newcommand{\NumAblEarliestSparseStagnations}{0}
\newcommand{\NumAblFdDenseSolveRate}{100}
\newcommand{\NumAblBaselineDenseSolveRate}{100}
\newcommand{\NumAblFdDense}{12.48}
\newcommand{\NumAblDenseEvaluations}{28}
\newcommand{\NumAblFdDefaultSolveRate}{100}
\newcommand{\NumAblBaselineDefaultSolveRate}{100}
\newcommand{\NumAblFdDefault}{1.97}
\newcommand{\NumAblFdSparseSolveRate}{100}
\newcommand{\NumAblBaselineSparseSolveRate}{100}
\newcommand{\NumAblFdSparse}{1.79}
\newcommand{\NumAblEarliestDenseSolveRate}{92}
\newcommand{\NumAblEarliestDense}{5.91}
\newcommand{\NumAblEarliestDefaultSolveRate}{100}
\newcommand{\NumAblEarliestDefault}{4.29}
\newcommand{\NumAblEarliestSparseSolveRate}{100}
\newcommand{\NumAblEarliestSparse}{3.89}
\newcommand{\NumAblSciSeverityDenseSolveRate}{98}
\newcommand{\NumAblSciSeverityDense}{12.3}
\newcommand{\NumAblOrderMilliseconds}{0}
\newcommand{\NumAblSciIndexDenseSolveRate}{96}
\newcommand{\NumAblSciIndexDense}{14.5}
\newcommand{\NumAblSciDenseSolveRateMin}{96}
\newcommand{\NumAblSciDenseSolveRateMax}{100}
\newcommand{\NumAblScpSeverityDenseSolveRate}{72}
\newcommand{\NumAblScpSeverityDense}{3.5}
\newcommand{\NumAblScpIndexDenseSolveRate}{56}
\newcommand{\NumAblScpIndexDense}{11.1}
\newcommand{\NumAblScpDenseSolveRateMin}{56}
\newcommand{\NumAblScpDenseSolveRateMax}{72}
\newcommand{\NumAblScpViolationDenseMin}{0.14}
\newcommand{\NumAblScpViolationDenseMax}{0.19}
\newcommand{\NumAblSciSeverityDefaultSolveRate}{100}
\newcommand{\NumAblSciIndexDefaultSolveRate}{100}
\newcommand{\NumAblSciDefaultSolveRateMin}{100}
\newcommand{\NumAblSciDefaultSolveRateMax}{100}
\newcommand{\NumAblScpSeverityDefaultSolveRate}{90}
\newcommand{\NumAblScpIndexDefaultSolveRate}{80}
\newcommand{\NumAblScpDefaultSolveRateMin}{80}
\newcommand{\NumAblScpDefaultSolveRateMax}{90}
\newcommand{\NumAblScpViolationDefaultMin}{0.45}
\newcommand{\NumAblScpViolationDefaultMax}{0.81}
\newcommand{\NumFleetSuccessRateSCI}{100}
\newcommand{\NumFleetSuccessRateSCPLast}{98}
\newcommand{\NumFleetContinuousRateSCPLast}{0}
\newcommand{\NumAccSuccessRateSCI}{100}
\newcommand{\NumAblBaselineSuccessRate}{100}
\newcommand{\NumAblBaselineDenseSuccessRate}{100}
\newcommand{\NumAblBaselineDefaultSuccessRate}{100}
\newcommand{\NumAblBaselineSparseSuccessRate}{100}
\newcommand{\NumAblFdSuccessRate}{100}
\newcommand{\NumAblFdDenseSuccessRate}{100}
\newcommand{\NumAblFdDefaultSuccessRate}{100}
\newcommand{\NumAblFdSparseSuccessRate}{100}
\newcommand{\NumAblEarliestSuccessRate}{97.3}
\newcommand{\NumAblEarliestDenseSuccessRate}{92}
\newcommand{\NumAblEarliestDefaultSuccessRate}{100}
\newcommand{\NumAblEarliestSparseSuccessRate}{100}
\newcommand{\NumPrioSuccessRateSCI}{100} 
\newcommand{\NumMatchedSnapshot}{2026-09-23, 15:26 UTC}
\newcommand{\NumMatchedCompleted}{800}
\newcommand{\NumMatchedRemaining}{0}
\newcommand{\NumMatchedNFiveDTwoSCIAccepted}{50}
\newcommand{\NumMatchedNFiveDTwoSCICompleted}{50}
\newcommand{\NumMatchedNFiveDTwoSCIValid}{50}
\newcommand{\NumMatchedNFiveDTwoSCIAttempts}{50}
\newcommand{\NumMatchedNFiveDTwoSCITimeouts}{0}
\newcommand{\NumMatchedNFiveDTwoSCICappedMedian}{0.000294}
\newcommand{\NumMatchedNFiveDTwoSCIMedian}{0.000294}
\newcommand{\NumMatchedNFiveDTwoSCIMean}{0.00102}
\newcommand{\NumMatchedNFiveDTwoSCPAccepted}{43}
\newcommand{\NumMatchedNFiveDTwoSCPCompleted}{50}
\newcommand{\NumMatchedNFiveDTwoSCPValid}{50}
\newcommand{\NumMatchedNFiveDTwoSCPAttempts}{50}
\newcommand{\NumMatchedNFiveDTwoSCPTimeouts}{0}
\newcommand{\NumMatchedNFiveDTwoSCPCappedMedian}{0.148}
\newcommand{\NumMatchedNFiveDTwoSCPMedian}{0.148}
\newcommand{\NumMatchedNFiveDTwoSCPMean}{1}
\newcommand{\NumMatchedNFiveDTwoMedianRuntimeRatio}{504}
\newcommand{\NumMatchedNFiveDThreeSCIAccepted}{50}
\newcommand{\NumMatchedNFiveDThreeSCICompleted}{50}
\newcommand{\NumMatchedNFiveDThreeSCIValid}{50}
\newcommand{\NumMatchedNFiveDThreeSCIAttempts}{50}
\newcommand{\NumMatchedNFiveDThreeSCITimeouts}{0}
\newcommand{\NumMatchedNFiveDThreeSCICappedMedian}{0.000319}
\newcommand{\NumMatchedNFiveDThreeSCIMedian}{0.000319}
\newcommand{\NumMatchedNFiveDThreeSCIMean}{0.000982}
\newcommand{\NumMatchedNFiveDThreeSCPAccepted}{49}
\newcommand{\NumMatchedNFiveDThreeSCPCompleted}{50}
\newcommand{\NumMatchedNFiveDThreeSCPValid}{49}
\newcommand{\NumMatchedNFiveDThreeSCPAttempts}{50}
\newcommand{\NumMatchedNFiveDThreeSCPTimeouts}{0}
\newcommand{\NumMatchedNFiveDThreeSCPCappedMedian}{0.149}
\newcommand{\NumMatchedNFiveDThreeSCPMedian}{0.149}
\newcommand{\NumMatchedNFiveDThreeSCPMean}{1.72}
\newcommand{\NumMatchedNFiveDThreeMedianRuntimeRatio}{467}
\newcommand{\NumMatchedNFiveDFiveSCIAccepted}{50}
\newcommand{\NumMatchedNFiveDFiveSCICompleted}{50}
\newcommand{\NumMatchedNFiveDFiveSCIValid}{50}
\newcommand{\NumMatchedNFiveDFiveSCIAttempts}{50}
\newcommand{\NumMatchedNFiveDFiveSCITimeouts}{0}
\newcommand{\NumMatchedNFiveDFiveSCICappedMedian}{0.00092}
\newcommand{\NumMatchedNFiveDFiveSCIMedian}{0.00092}
\newcommand{\NumMatchedNFiveDFiveSCIMean}{0.00142}
\newcommand{\NumMatchedNFiveDFiveSCPAccepted}{48}
\newcommand{\NumMatchedNFiveDFiveSCPCompleted}{50}
\newcommand{\NumMatchedNFiveDFiveSCPValid}{48}
\newcommand{\NumMatchedNFiveDFiveSCPAttempts}{50}
\newcommand{\NumMatchedNFiveDFiveSCPTimeouts}{0}
\newcommand{\NumMatchedNFiveDFiveSCPCappedMedian}{0.168}
\newcommand{\NumMatchedNFiveDFiveSCPMedian}{0.168}
\newcommand{\NumMatchedNFiveDFiveSCPMean}{1.93}
\newcommand{\NumMatchedNFiveDFiveMedianRuntimeRatio}{182}
\newcommand{\NumMatchedNFiveDTenSCIAccepted}{50}
\newcommand{\NumMatchedNFiveDTenSCICompleted}{50}
\newcommand{\NumMatchedNFiveDTenSCIValid}{50}
\newcommand{\NumMatchedNFiveDTenSCIAttempts}{50}
\newcommand{\NumMatchedNFiveDTenSCITimeouts}{0}
\newcommand{\NumMatchedNFiveDTenSCICappedMedian}{0.000286}
\newcommand{\NumMatchedNFiveDTenSCIMedian}{0.000286}
\newcommand{\NumMatchedNFiveDTenSCIMean}{0.000865}
\newcommand{\NumMatchedNFiveDTenSCPAccepted}{49}
\newcommand{\NumMatchedNFiveDTenSCPCompleted}{50}
\newcommand{\NumMatchedNFiveDTenSCPValid}{49}
\newcommand{\NumMatchedNFiveDTenSCPAttempts}{50}
\newcommand{\NumMatchedNFiveDTenSCPTimeouts}{0}
\newcommand{\NumMatchedNFiveDTenSCPCappedMedian}{0.209}
\newcommand{\NumMatchedNFiveDTenSCPMedian}{0.209}
\newcommand{\NumMatchedNFiveDTenSCPMean}{1.14}
\newcommand{\NumMatchedNFiveDTenMedianRuntimeRatio}{733}
\newcommand{\NumMatchedNTenDTwoSCIAccepted}{50}
\newcommand{\NumMatchedNTenDTwoSCICompleted}{50}
\newcommand{\NumMatchedNTenDTwoSCIValid}{50}
\newcommand{\NumMatchedNTenDTwoSCIAttempts}{50}
\newcommand{\NumMatchedNTenDTwoSCITimeouts}{0}
\newcommand{\NumMatchedNTenDTwoSCICappedMedian}{0.00308}
\newcommand{\NumMatchedNTenDTwoSCIMedian}{0.00308}
\newcommand{\NumMatchedNTenDTwoSCIMean}{0.00428}
\newcommand{\NumMatchedNTenDTwoSCPAccepted}{13}
\newcommand{\NumMatchedNTenDTwoSCPCompleted}{50}
\newcommand{\NumMatchedNTenDTwoSCPValid}{48}
\newcommand{\NumMatchedNTenDTwoSCPAttempts}{50}
\newcommand{\NumMatchedNTenDTwoSCPTimeouts}{0}
\newcommand{\NumMatchedNTenDTwoSCPCappedMedian}{8.43}
\newcommand{\NumMatchedNTenDTwoSCPMedian}{8.43}
\newcommand{\NumMatchedNTenDTwoSCPMean}{12.4}
\newcommand{\NumMatchedNTenDTwoMedianRuntimeRatio}{2734}
\newcommand{\NumMatchedNTenDThreeSCIAccepted}{50}
\newcommand{\NumMatchedNTenDThreeSCICompleted}{50}
\newcommand{\NumMatchedNTenDThreeSCIValid}{50}
\newcommand{\NumMatchedNTenDThreeSCIAttempts}{50}
\newcommand{\NumMatchedNTenDThreeSCITimeouts}{0}
\newcommand{\NumMatchedNTenDThreeSCICappedMedian}{0.0035}
\newcommand{\NumMatchedNTenDThreeSCIMedian}{0.0035}
\newcommand{\NumMatchedNTenDThreeSCIMean}{0.00481}
\newcommand{\NumMatchedNTenDThreeSCPAccepted}{50}
\newcommand{\NumMatchedNTenDThreeSCPCompleted}{50}
\newcommand{\NumMatchedNTenDThreeSCPValid}{50}
\newcommand{\NumMatchedNTenDThreeSCPAttempts}{50}
\newcommand{\NumMatchedNTenDThreeSCPTimeouts}{0}
\newcommand{\NumMatchedNTenDThreeSCPCappedMedian}{15.1}
\newcommand{\NumMatchedNTenDThreeSCPMedian}{15.1}
\newcommand{\NumMatchedNTenDThreeSCPMean}{25.9}
\newcommand{\NumMatchedNTenDThreeMedianRuntimeRatio}{4308}
\newcommand{\NumMatchedNTenDFiveSCIAccepted}{50}
\newcommand{\NumMatchedNTenDFiveSCICompleted}{50}
\newcommand{\NumMatchedNTenDFiveSCIValid}{50}
\newcommand{\NumMatchedNTenDFiveSCIAttempts}{50}
\newcommand{\NumMatchedNTenDFiveSCITimeouts}{0}
\newcommand{\NumMatchedNTenDFiveSCICappedMedian}{0.00254}
\newcommand{\NumMatchedNTenDFiveSCIMedian}{0.00254}
\newcommand{\NumMatchedNTenDFiveSCIMean}{0.0038}
\newcommand{\NumMatchedNTenDFiveSCPAccepted}{49}
\newcommand{\NumMatchedNTenDFiveSCPCompleted}{50}
\newcommand{\NumMatchedNTenDFiveSCPValid}{49}
\newcommand{\NumMatchedNTenDFiveSCPAttempts}{50}
\newcommand{\NumMatchedNTenDFiveSCPTimeouts}{0}
\newcommand{\NumMatchedNTenDFiveSCPCappedMedian}{69.7}
\newcommand{\NumMatchedNTenDFiveSCPMedian}{69.7}
\newcommand{\NumMatchedNTenDFiveSCPMean}{75.7}
\newcommand{\NumMatchedNTenDFiveMedianRuntimeRatio}{27397}
\newcommand{\NumMatchedNTenDTenSCIAccepted}{50}
\newcommand{\NumMatchedNTenDTenSCICompleted}{50}
\newcommand{\NumMatchedNTenDTenSCIValid}{50}
\newcommand{\NumMatchedNTenDTenSCIAttempts}{50}
\newcommand{\NumMatchedNTenDTenSCITimeouts}{0}
\newcommand{\NumMatchedNTenDTenSCICappedMedian}{0.00282}
\newcommand{\NumMatchedNTenDTenSCIMedian}{0.00282}
\newcommand{\NumMatchedNTenDTenSCIMean}{0.00812}
\newcommand{\NumMatchedNTenDTenSCPAccepted}{13}
\newcommand{\NumMatchedNTenDTenSCPCompleted}{50}
\newcommand{\NumMatchedNTenDTenSCPValid}{13}
\newcommand{\NumMatchedNTenDTenSCPAttempts}{50}
\newcommand{\NumMatchedNTenDTenSCPTimeouts}{37}
\newcommand{\NumMatchedNTenDTenSCPCappedMedian}{310}
\newcommand{\NumMatchedNTenDTenSCPMedian}{0.698}
\newcommand{\NumMatchedNTenDTenSCPMean}{52.1}
\newcommand{\NumMatchedNFiveSCIMedianMin}{0.000286}
\newcommand{\NumMatchedNFiveSCIMedianMax}{0.00092}
\newcommand{\NumMatchedNFiveSCIContactMedianMin}{0}
\newcommand{\NumMatchedNFiveSCIContactMedianMax}{0.5}
\newcommand{\NumMatchedNFiveSCIAccepted}{200}
\newcommand{\NumMatchedNFiveSCIAttempts}{200}
\newcommand{\NumMatchedNTenSCIMedianMin}{0.00254}
\newcommand{\NumMatchedNTenSCIMedianMax}{0.0035}
\newcommand{\NumMatchedNTenSCIContactMedianMin}{1}
\newcommand{\NumMatchedNTenSCIContactMedianMax}{3}
\newcommand{\NumMatchedNTenSCIAccepted}{200}
\newcommand{\NumMatchedNTenSCIAttempts}{200}
\newcommand{\NumMatchedDTwoSCIAccepted}{100}
\newcommand{\NumMatchedDTwoSCIAttempts}{100}
\newcommand{\NumMatchedDTwoSCISuccessRate}{100}
\newcommand{\NumMatchedDThreeSCIAccepted}{100}
\newcommand{\NumMatchedDThreeSCIAttempts}{100}
\newcommand{\NumMatchedDThreeSCISuccessRate}{100}
\newcommand{\NumMatchedDFiveSCIAccepted}{100}
\newcommand{\NumMatchedDFiveSCIAttempts}{100}
\newcommand{\NumMatchedDFiveSCISuccessRate}{100}
\newcommand{\NumMatchedDTenSCIAccepted}{100}
\newcommand{\NumMatchedDTenSCIAttempts}{100}
\newcommand{\NumMatchedDTenSCISuccessRate}{100}
\newcommand{\NumMatchedFinalCell}{At $N=10,d=10$, median runtimes are $0.00282\,\mathrm{s}$ for \gls{sci} and $0.698\,\mathrm{s}$ for \gls{scp}, with $50/50$ and $13/50$ accepted trajectories. Mean \gls{sci} runtime is $0.00812\,\mathrm{s}$. The other $37$ \gls{scp} attempts time out; its runtime median describes the surviving trajectories, including $7$ initially noninteracting instances.} 

\begin{figure}[t]
  \centering
  \includegraphics[width=\columnwidth]{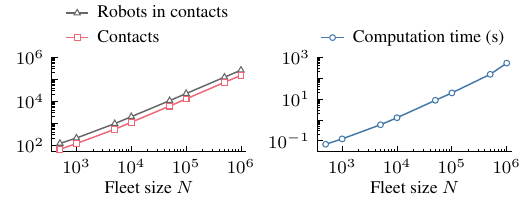}
  \caption{Grid-to-double-helix scaling (\cref{exp:million-robots}):
    contact counts, robots in contacts, and computation time versus fleet size,
    with one run per size.}
  \label{fig:numerics-helix}
\end{figure}

\begin{figure}[t]
  \centering
  \raisebox{0sp}[5815809sp][0sp]{\includegraphics[width=16516507sp,height=5815809sp]{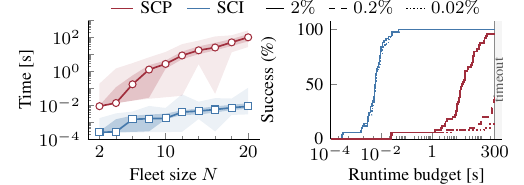}}
  \caption{Runtime and separation accuracy.
    Left: computation time for solver-converged runs over $50$ seeds per fleet
    size (\cref{exp:fleet-size}).
    Right: continuous-time separation success rates over $50$ instances at
    $N=16$ versus runtime budget (\cref{exp:accuracy}).
    Colors identify methods in both panels; line styles on the right identify
    separation targets relative to $R$.}
  \label{fig:numerics-comparison}
\end{figure}

Unless stated otherwise, the experiments use planar fleets with $R=0.5$
and zero initial and final velocities. We sample initial and final configurations
independently in a disk of radius $5\sqrt{N/2}$, rejecting configurations
with pairwise distances below $0.8$. The horizon is $T=10\sqrt{N/2}$.
The measurements were obtained on an Apple M3 Pro with 36\,GB of
memory. We use the current \gls{sci} implementation and
\gls{scp} measurements on the same instances.
The scaling and baseline comparisons use unrestricted boxes;
\cref{exp:box-bounds} tests finite position, velocity, and acceleration limits.

The baseline transcribes the same fixed-horizon problem with prescribed
terminal states, following~\citet[Section~III-C]{kamm_floating_2026}, with piecewise-constant
accelerations on $K=200$ equal intervals and OSQP
\citep{stellato_osqp_2020} for the quadratic subproblems.
In these comparisons, both methods start from the unconstrained solution.
Unless varied below,
$\varepsilon_R=10^{-3}$, or $0.2\%$ of $R$.
For cubic contact matching, $\varepsilon_F=10^{-7}$,
$\varepsilon_m=10^{-9}$, and the Newton iteration limit is $100$.
The baseline uses at most $20$ outer iterations with position-change
tolerance $10^{-2}$; OSQP uses absolute tolerance $10^{-4}$, zero
relative tolerance, and at most $10^4$ iterations. The comparison studies
use a $300\,\mathrm{s}$ time limit.

Success requires solver convergence, satisfaction of the imposed box bounds,
and pairwise separation of at least
$R-\varepsilon_R$: throughout $[0,T]$ for \gls{sci}, and at the enforced
grid points for \gls{scp}.
The \emph{continuous-time separation success rate} counts successful runs
whose piecewise-polynomial trajectories also meet this bound throughout
$[0,T]$. Both rates use all attempted instances, including failures and
timeouts, in their denominator; these values coincide for \gls{sci}.
In plots with shaded bands, lines show medians; darker and lighter bands
show the interquartile and central $95\%$ ranges across seeds.

\begin{experiment}[A million robots]
\label{exp:million-robots}
We compute transitions from a planar grid to a double helix for fleets
of $500$ to $1{,}000{,}000$ robots (\cref{fig:header,fig:numerics-helix}).
The workspace grows with fleet size while the initial grid spacing
remains $2$\,m. We use a horizon of $100$\,s and a separation tolerance
of $1$\,cm.
\emph{Results.}
For $1{,}000$ robots, we compute trajectories with $118$ contacts
in $0.124$\,s. For $10^6$ robots, we obtain $151{,}071$ contacts
in $525.4$\,s: we plan continuous-time motion for a million robots in
under nine minutes on a single workstation.
Sparse interactions keep the largest final contact component to only $16$
robots, allowing us to represent the million-robot solution through small
coupled systems.
\end{experiment}

\begin{experiment}[Fleet size]
\label{exp:fleet-size}
We study the effect of fleet size on runtime and success rate, using
$50$ seeds for each $N\in\{2,4,\ldots,20\}$.
\emph{Results.}
\Gls{sci} has a $\NumFleetSuccessRateSCI\%$ success rate across
$\NumFleetAttemptsSCI$ instances. At $N=20$, its median runtime is
$\NumFleetSCI\,\mathrm{s}$ (\cref{fig:numerics-comparison}, left), with medians of
$\NumFleetContactMedian$ final contacts and $\NumFleetSystemSizeMedian$
matching unknowns across the contact components.
At $N=20$, the baseline has a $\NumFleetSuccessRateSCPLast\%$ success rate
and median runtime $\NumFleetSCP\,\mathrm{s}$; its continuous-time
separation success rate is $\NumFleetContinuousRateSCPLast\%$.
Its median continuous-time separation deficit is
$\NumFleetEtaSCPLast\%$ of $R$.
At $N=20$, \gls{sci} computes trajectories $\NumFleetRatio$ times faster
than \gls{scp}.
The baseline's high convergence rate on the fixed grid therefore
substantially overstates continuous-time feasibility at this resolution.
\end{experiment}

\begin{experiment}[Separation accuracy and cost]
\label{exp:accuracy}
For $50$ instances at $N=16$, we set
$\varepsilon_R\in\{10^{-2},10^{-3},10^{-4}\}$, corresponding to
$2\%$, $0.2\%$, and $0.02\%$ of $R$.
The baseline is run at $K\in\{100,200,400,800,1600\}$ with OSQP
absolute tolerance $\varepsilon_R/10$.
For each instance and target, the baseline curve uses the fastest
successful run that meets the continuous-time separation target.
\emph{Results.}
Across a hundredfold tightening of the separation tolerance,
\gls{sci} retains a $\NumAccSuccessRateSCI\%$ success rate and a nearly unchanged
median runtime of $\NumAccSCIRuntimeMax\,\mathrm{s}$, with
a median contact count of $\NumAccContactMedianRange$.
Isolated contacts satisfy the same matching equations at every separation
tolerance.
By contrast, \gls{scp} requires finer time grids, and its continuous-time
separation success rate falls from
$\NumAccContinuousRateLoose\%$ to $\NumAccContinuousRateMedium\%$ and
$\NumAccContinuousRateStrict\%$ at the three targets
(\cref{fig:numerics-comparison}, right).
At the $2\%$ target, the median speedup on the $\NumAccInteractingLoose$
interacting instances where both methods meet continuous-time separation
is $\NumAccSpeedupLoose$.
At $0.2\%$ and $0.02\%$, median speedups are
$\NumAccSpeedupMedium$ ($\NumAccInteractingMedium$ instances) and
$\NumAccSpeedupStrict$ ($\NumAccInteractingStrict$ instances).
Against the coarsest successful baseline grid at each target, the maximum
relative cost difference across $\NumAccCostPairs$ paired successes is
$\NumAccCostMaxDifference\%$.
The large runtime gains therefore accompany nearly identical acceleration
costs on these paired feasible instances.
\end{experiment}

\begin{experiment}[Congestion]
\label{exp:congestion}
At fixed $N=10$ and $T=10\sqrt{N/2}$, we vary
$s_{\mathrm{ws}}\in\{0.12,\allowbreak 0.15,\allowbreak 0.175,\allowbreak
0.4,\allowbreak 0.55,\allowbreak 0.75,\allowbreak 1,\allowbreak
1.25,\allowbreak 1.5\}$,
with $50$ instances per scale, spanning sparse to extreme congestion.
At $s_{\mathrm{ws}}=0.12$, the endpoint exclusion disks cover $52.7\%$
of their enclosing disk, close to the bulk random sequential adsorption
(RSA) jamming coverage of $54.7\%$.
The nominal physical robot-area fraction is $34.7\%$.
Accepted endpoint pairs at $0.12$ are also scaled to $0.15$ and $0.175$.
\emph{Results.}
Both methods solve all $150$ instances at the three densest levels.
At $s_{\mathrm{ws}}=0.12$, all $10$ robots enter a single matching component
in $15/50$ instances (percolation).
Median \gls{sci} runtime is $0.20\,\mathrm{s}$, with a maximum of
$0.65\,\mathrm{s}$, compared with a median of $22.87\,\mathrm{s}$
for \gls{scp}.
\Gls{sci} therefore retains subsecond runtimes even when the fleet cannot
be split into smaller independent contact components.
\end{experiment}

\begin{experiment}[Algorithmic ablations]
\label{exp:ablations}
Using the same $50$ seeds at $N=10$ and
$s_{\mathrm{ws}}\in\{0.4,1,1.5\}$, we replace the analytical Jacobian
with finite differences, or insert at the earliest violating local
distance minimum instead of the worst violation.
\emph{Results.}
At $s_{\mathrm{ws}}=0.4$, finite differences increase runtime by a median
factor of $\NumAblFdDense$ on jointly successful instances.
Earliest insertion stagnates on $\NumAblEarliestDenseStagnations$ of the
$\NumAblBaselineDenseAttempts$ dense instances; worst-violation insertion
solves all $\NumAblBaselineDenseAccepted$.
Analytical derivatives therefore improve speed, while worst-violation
insertion avoids the observed stagnation.
\end{experiment}

\begin{samepage}
\begin{experiment}[Spatial dimension]
\label{exp:dimension}
We test $d\in\{2,3,5,10\}$ and $N\in\{5,10\}$ with $T=10$.
Endpoints are sampled in a $d$-dimensional ball, whose radius is
calibrated on $256$ separate seeds to match the planar mean number of
unconstrained interacting pairs. Each $(d, N)$ is evaluated on
$50$ further seeds.
\emph{Results.}
\Gls{sci} succeeds on all instances. At $N=10$, its median runtime stays
within $\NumMatchedNTenSCIMedianMin$--$\NumMatchedNTenSCIMedianMax\,\mathrm{s}$
across dimensions.
Its speedup over \gls{scp} increases with dimension
($d=2$: $\NumMatchedNTenDTwoMedianRuntimeRatio\times$,
$d=3$: $\NumMatchedNTenDThreeMedianRuntimeRatio\times$,
$d=5$: $\NumMatchedNTenDFiveMedianRuntimeRatio\times$), using median runtimes.
At $d=10$, \gls{scp} times out on
$\NumMatchedNTenDTenSCPTimeouts$ out of $\NumMatchedNTenDTenSCPAttempts$ instances.
The stable runtimes show that contact matching remains practical in higher
dimensions when encounter frequency is controlled.
\end{experiment}
\end{samepage}

\begin{experiment}[Box bounds]
\label{exp:box-bounds}
We solve \eqref{eq:fleet-problem} with \gls{sci} on $102$ instances
with $N=3$--$12$ and feasible \gls{scp} reference trajectories.
The suite contains $68$ planar and $34$ spatial cases: crossing chains,
circle swaps, and corridors, with $T=6.5$--$14.9\,\mathrm{s}$ and
$R\in\{0.30,0.42,0.50\}\,\mathrm{m}$.
In $101$ cases, the componentwise limits are symmetric:
$(\overline b^{(1)},\overline b^{(2)})=(0.8,0.65)$ in $74$ cases,
$(0.65,0.4)$ in $24$, and $(0.75,0.55)$ in $3$, in
$\mathrm{m/s}$ and $\mathrm{m/s^2}$, respectively.
One case uses asymmetric robot- and axis-dependent bounds.
The $23$ corridor cases also impose $|x_{1,2}(t)|\leq0.04\,\mathrm{m}$;
all other position coordinates are unrestricted.
\emph{Results.}
All $102$ bounded instances pass continuous feasibility checks
at tolerance $2\times10^{-6}$.
At this tolerance, position, velocity, and acceleration limits are active
in $23$, $85$, and $58$ cases, respectively.
The median runtime across the bounded instances is $0.20\,\mathrm{s}$.
\end{experiment} 
\section{Conclusion}
\label{sec:conclusion}

By optimizing sparse contact components, \gls{sci} makes million-agent
planning computationally feasible while retaining continuous-time
separation checks.
An extension with fixed virtual robots could encode static obstacles as
unions of balls, enforcing separation only for pairs involving a physical
robot. This would let large agent counts represent environmental detail
as well as physical fleets. 

\ifnum\RSSAnonymous=1
  \clearpage
\fi
\begingroup
\def\doi#1.{\unskip}

\endgroup
\end{document}